\documentclass[11pt]{article}

\usepackage[final]{acl}
 \usepackage{booktabs} 
\usepackage{times}
\usepackage{latexsym}
\usepackage{multirow} 
\usepackage[T1]{fontenc}

\usepackage[utf8]{inputenc}

\usepackage{microtype}

\usepackage{inconsolata}

\usepackage{graphicx}

\title{Designing an Auditable LLM-Supported Workflow for Qualitative Thematic Analysis}

\author{
  \textbf{Nadia Jul Jeldtoft}\textsuperscript{1}
  \qquad
  \textbf{Tariq Yousef}\textsuperscript{2}
  \\
  \textsuperscript{1}Department of Culture and Language \\
  \textsuperscript{2}Department of Mathematics and Computer Science \\
  University of Southern Denmark \\
  \texttt{najel@sdu.dk}
  \qquad
  \texttt{yousef@imada.sdu.dk}
}

\begin{document}
\maketitle
\begin{abstract}
Large Language Models (LLMs) offer new possibilities for scaling qualitative analysis, but existing applications often provide limited methodological transparency regarding how qualitative methods are translated into computational procedures. This paper presents an auditable and privacy-preserving computational operationalization of inductive and latent Thematic Analysis (TA). This paper first derives five design principles from the methodological requirements of TA and the conditions introduced by LLM-based inference: preserving interpretative context, maintaining traceable relationships between empirical material and analytical outputs, representing analytical constructs and reasoning explicitly, constraining LLM inference to interpretative tasks, and enabling privacy-preserving local deployment. Second, it presents a proof-of-concept for a two-phase workflow that operationalizes these principles by combining interpretative LLM inference with deterministic procedural control to generate codes, analytical justifications, themes, and theme descriptions while preserving explicit links to the source material. Third, it proposes an evaluation framework combining structural comparison with human-led TA and independent expert assessment of analytical quality. The evaluation is conducted on semi-structured Danish interview transcripts. and the results shows that the workflow produces code-level outputs with coverage broadly comparable to human annotations and highly rated analytical justifications, while generating a more compressed thematic structure characterized by fewer and broader themes. The findings demonstrate the feasibility of auditable LLM-supported TA through a modular workflow designed to scale to larger datasets, accommodate different LLMs, and support transfer across research domains, with domain adaptation primarily requiring adjustments to the prompting strategy.
\end{abstract}

\section{Introduction}

Large Language Models (LLMs) have expanded the possibilities for analyzing large collections of unstructured textual data. Their ability to process textual context, identify patterns and generate human-interpretable analytical outputs has led to their increasing application in qualitative research \cite{barros2025llmqualitative, davison_ethics_2024}. 
However, qualitative analysis is defined not only by its analytical outputs but by the methodological principles through which analytical outputs are generated. Across qualitative research traditions, these principles include contextual interpretation and iterative engagement with empirical material\cite{kvale_interviews_2015, hastrup_getting_2004, denzin2011sage}. 

\textsc{Thematic Analysis}  (TA) \cite{braun_using_2006, boyatzis_transforming_1998} is a foundational method for the analysis of qualitative data across disciplines. It provides a systematic, yet flexible framework for identifying and interpreting patterns of meaning in diverse research contexts \cite {kvale_interviews_2015, denzin2011sage}. TA can be understood as an interpretative process that, in part, resembles unsupervised learning tasks such as pattern detection and clustering in unlabeled text data \cite{tan_introduction_2019}. However, unlike purely computational approaches, TA is guided by meaning and depends on the analyst’s theoretical and epistemological perspective. In practice, TA involves iteratively assigning codes to meaningful features of the data and grouping these into higher-level themes while maintaining explicit links between empirical data, codes and themes. Consequently, computational operationalizations of TA must support not only code and theme generation, but also contextual interpretation, analytical reasoning and traceable relationships across the analytical process. 

Existing LLM-supported approaches to TA  primarily demonstrate the analytical capabilities of the models themselves, while providing  limited methodological accounts of how qualitative analytical methods are operationalized computationally. 
The challenge is therefore not \textit{whether} LLMs can generate codes and themes, but \textit{how }LLM inference can be utilized in computational workflows that preserve the methodological principles of qualitative analysis. Without careful workflow design, the use of LLMs risks obscuring rather than supporting the interpretative analytical process central to qualitative analysis \cite {beltoft_not_2025, davison_ethics_2024}. 

This paper presents a a proof of concept for an auditable computational operationalization of TA that translates the method into a structured computational workflow using an LLM for analysis. In this context, auditability refers to the ability to systematically document and critically assess how analytical outputs are generated, supported by transparency across intermediate analytical operations and traceability between outputs and their empirical source material \cite {kroll2021traceability, liao2024transparency, mora_cantallops_traceability_2021}.

The workflow separates interpretative inference from deterministic procedural computation. It embeds methodological quality criteria from TA directly into the computational design, and preserves explicit links between data, codes and themes throughout the analytical process in a privacy-preserving manner \cite {srinivasan_start_2018}. The proposed workflow is evaluated through structural comparison with expert-generated TA and independent expert assessment on a corpus of semi-structured Danish interview transcripts. 

This paper makes three contributions: First, it proposes five design principles for the computational operationalization of TA grounded in the methodological principles of qualitative research. Second, it introduces an auditable workflow architecture that combines constrained LLM inference with deterministic programmatic procedures to preserve analytical transparency, traceability, and privacy throughout the analytical process. Third, it proposes an evaluation framework for assessing whether computational operationalizations of TA preserve central methodological quality criteria through complementary structural comparison and expert assessment

\section{Related work}

Although recent studies demonstrate the functional feasibility of LLMs for supporting TA, they provide  limited methodological descriptions of the computational architecture, analytical representations and workflow design decisions that translate core  principles from qualitative methodology into computational procedures.
This is reflected in limited descriptions of prompt design, data representations, coding procedures and theme construction, making it difficult to assess how analytical outputs are generated and how methodological principles are preserved throughout the workflow \cite {perkins2024use, ornelas2025llm}. 

More recent studies  \cite {de_paoli_performing_2024, montes_large_2025} have begun to introduce multi-stage workflows and methodologically informed prompting strategies. 
Nevertheless, key workflow design decisions remain underspecified, including how analytical units are represented, how contextual information is managed across inferential stages, how intermediate analytical products are linked, and how analytical traceability is maintained throughout the workflow. 

A further limitation is that many existing solutions rely on proprietary cloud-based models and API infrastructures. Besides limiting methodological transparency, this raises challenges for governance and the analysis of sensitive qualitative data. 

Taken together, existing work demonstrates that LLMs can support core analytical tasks in TA. However, comparatively little attention has been given to the computational operationalization of the method itself. This leaves a need for workflow architectures that explicitly embed the methodological principles of TA into the computational design while preserving analytical traceability, transparency and data governance.

\section{Background Theory}
In this work, we adopt the framework for TA proposed by \citet{braun_using_2006}. TA provides a flexible methodological framework in which themes are developed through iterative interpretation of meaningful patterns in the data. Because TA depend on the analytical focus and epistemological perspective of the researcher, there is no single "correct" thematic representation independent of the analytical context and research focus. A central characteristic of TA is that pattern identification is guided by meaning rather than formal similarity. 

Consequently, the identification and grouping of codes into themes depend on contextual interpretation and the focus of the analyst rather than predefined computational rules. In this work, we adopt an inductive and latent approach in which codes and themes are iteratively derived from underlying patterns of meaning in the data \cite{braun_using_2006}. 

The workflow for conducting TA consists of two overall analytical phases. The first phase (from text to codes) involves systematic coding of the dataset, where interview segments are assigned one or more codes. The second phase (from codes to themes) involves constructing broader themes by identifying patterns of shared meaning across codes. 

\citet{braun_using_2006} outline a set of quality criteria for conducting TA that extend beyond the analytical outputs themselves. Based on these quality criteria, we derive four methodological dimensions of TA: the analytical process, the representation of analytical constructs, the traceability of analytical relationships, and the characteristics of analytically valid outputs. Table~\ref{tab:methodological_dimensions} summarizes these methodological dimensions. 

\begin{table*}[t]
\centering
\footnotesize

\caption{Methodological dimensions underlying the quality criteria of Thematic Analysis, adapted from Braun \& Clarke}
\label{tab:methodological_dimensions}

\begin{tabular}{p{0.65\textwidth} p{0.29\textwidth}}
\toprule

\textbf{TA quality criterion} &
\textbf{Methodological dimension} \\

\midrule

Systematic and thorough coding across the entire dataset &
Analytical process \\[0.4em]

Themes organize multiple related codes into broader interpretative patterns &
Analytical process \\[0.4em]

Code and theme labels must provide clear and meaningful representations &
Analytical representation \\[0.4em]

Theme descriptions should describe the content and meaning of the theme &
Analytical representation \\[0.4em]

Transparent links between empirical data, codes and themes &
Traceability \\[0.4em]

Generated codes and themes should constitute analytically meaningful interpretations aligned with the analytical focus &
Analytical outputs \\

\bottomrule
\end{tabular}

\end{table*}

These  quality criteria and methodological dimensions  serve different functions in the computational operationalization of TA. The dimensions of analytical process, representation, and traceability specify central characteristics that must be preserved in the computational design, while the characteristics of analytically valid outputs provide criteria for evaluating the resulting analysis. These methodological considerations must furthermore be combined with the opportunities and limitations introduced by LLM-based analysis, including probabilistic inference and the governance of sensitive qualitative data. 

\section{Five design requirements for adapting LLMs for TA}

LLMs possess a dual capability that makes them particularly relevant for computational TA. Through their training on large-scale textual corpora, they are capable not only of generating text, but also of identifying and organizing patterns of meaning in unstructured textual data \cite{hao_reasoning_2023,floridi_ai_2023,naveed_comprehensive_2024} 
However,  their application is not methodologically neutral. Successfully operationalizing TA therefore requires more than applying an LLM to qualitative data: it requires explicit computational design decisions that preserve the methodological characteristics of TA while accommodating the opportunities and limitations of LLM-based inference. A computational operationalization must therefore account both for the methodological characteristics of TA and for the conditions introduced by LLM-based inference. Based on these two sources, we derive five design principles that guide the architecture of the proposed workflow.

\subsection{Principle 1: Preserve interpretative context}

TA relies on contextual interpretation, where codes are assigned to meaningful patterns within their local context   \cite {braun_using_2006}. Consequently, a computational operationalization of TA requires analytical units that preserve sufficient contextual information to support meaningful interpretation.

When utilizing an LLM for analytical purposes, this principle introduces an additional computational consideration. Analytical units must be sufficiently large to provide the contextual information required for interpretation, yet sufficiently small to support stable inference and consistent code generation within the model's context window \cite {liu_lost_2024}. Data segmentation therefore represents a methodological design decision rather than a preprocessing step, as the structure of the analytical units directly shapes the analytical outputs.

\subsection{Principle 2: Preserve traceable relationships between empirical data and analytical outputs }
The methodological dimension of traceability requires that analytical relationships remain explicit throughout the analytical process. In TA, this entails maintaining transparent links from empirical data to codes and from codes to themes. In a computational operationalization, codes, themes, and their associated justifications and descriptions must therefore be linked in a manner that enables the complete analytical path from themes, through codes, and back to the original data to be reconstructed in a deterministic manner.

\subsection{Principle 3: Represent analytical constructs and reasoning explicitly }
The methodological dimension of analytical representation requires that analytical constructs remain explicit and interpretable throughout the analytical process. In TA, this entails representing codes and themes through clear labels and descriptions, while making the analytical rationale for coding and theme construction transparent and grounded in the empirical material. When an LLM is used for analysis, these analytical constructs and their associated rationales must therefore be expressed as inspectable outputs rather than remaining implicit within model inference. 

\subsection{Principle 4: Constrain inference}

The methodological dimension of the analytical process requires contextual judgement when identifying and abstracting patterns of meaning. LLMs are well suited to support such tasks because they can process textual context and generate human-interpretable analytical outputs. However, these outputs are produced through probabilistic inference rather than deterministic procedures. In a computational operationalization of TA, LLM inference should therefore be limited to analytical tasks that require contextual interpretation, while procedural operations that do not require interpretation should be handled deterministically. This establishes \textit{a functional separation of labor} between\textit{ interpretative tasks} and the \textit{procedural tasks} of computational control.

\subsection{Principle 5: Preserve privacy though local deployment}
Qualitative interview data often contain sensitive information, making privacy preservation a methodological requirement. Cloud-based infrastructures may pose challenges in relation to  data governance and transparency. Consequently, computational operationalizations of TA should be designed to support privacy-preserving deployment \cite{srinivasan_start_2018}. This requires deployment strategies that support local inference using open-weight language models without exposing sensitive data to external services. 

\section{Data}
The data for the development of the workflow stems from the research project {\textit{The Family Revolution} (FREI) }\cite{bjerre_citizen_2022}. The project investigates micro-level experiences of family life and social change in Denmark between the 1950s and 1970s. The interviews provide a suitable case for developing and evaluating computational support for TA because they consist of rich, context-dependent qualitative narratives that require interpretative analysis.

The interviews were collected from 2021 to 2025 by Danish high school students. All interviews were transcribed with automated transcription software, followed by manual quality control and anonymization. A sample of 100 interview transcripts was selected from the larger dataset (n=570) using a stratified sampling strategy. First, interviews were stratified by length (long, medium, short) to reflect the distribution of the full dataset. Within each length category, interviews were then selected to achieve an approximately balanced gender distribution. Overall, this sample is intended to preserve key structural characteristics of the full dataset, including variation in interview length, narrative structure and thematic content. The sample dataset was randomly divided into a development and a test set at the interview level to separate model development from evaluation. After excluding one interview due to poor transcription quality, the final dataset consisted of 99 transcripts, where 5 transcripts were assigned to the test set and the remainder of 94 transcripts to the development set. 

\begin{figure*}[t]
\centering

\begin{minipage}{0.49\textwidth}
    \centering
    \includegraphics[width=\linewidth]{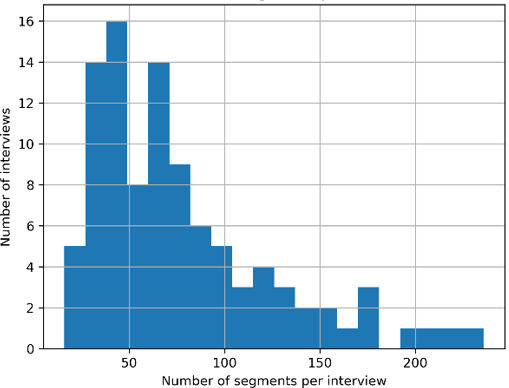}
    \caption{Distribution of segments per interview (n = 99)}
    \label{fig1}
\end{minipage}
\hfill
\begin{minipage}{0.49\textwidth}
    \centering
    \includegraphics[width=\linewidth]{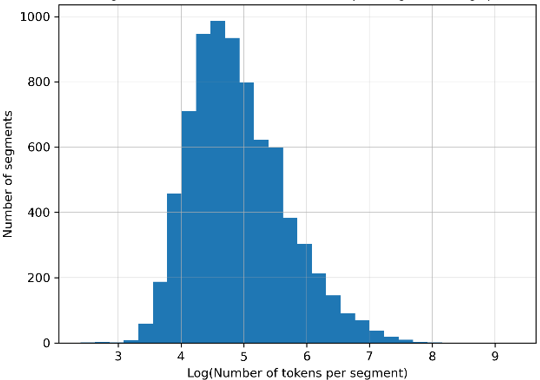}
    \caption{Log-transformed number of tokens per segment (n = 99)}
    \label{fig2}
\end{minipage}

\end{figure*}

To satisfy principle 1 (Preserve interpretable context), interview transcripts were segmented into analytical units consisting of two consecutive question-answer pairs (See Figures \ref{fig1} and \ref{fig2}). The choice of two consecutive question-answer pairs as the analytical unit represents a relevant level of segmentation for conducting TA. Smaller units risk loss of interpretative context and excessive overlap between generated codes, while larger units increase the risk of the LLM “losing” context by ignoring or overlooking parts of the input text when segments become too long. Initial experiments with different segment sizes suggested that longer input segments were associated with less stable outputs. This likely reflects a trade-off between prompt length , contextual complexity, and output generation within the model context window.  Although this relationship was not systematically evaluated, the selected segmentation strategy provides a sufficient basis for the development of the proof-of-concept. Exploration of optimal segment size remains a relevant area for future development and testing.

\section{Methodology}
\textbf {A functional division of labor} structures the workflow by aligning computational roles with the methodological requirements of TA.

\begin{figure*}[t]
\centering
\includegraphics[width=\textwidth]{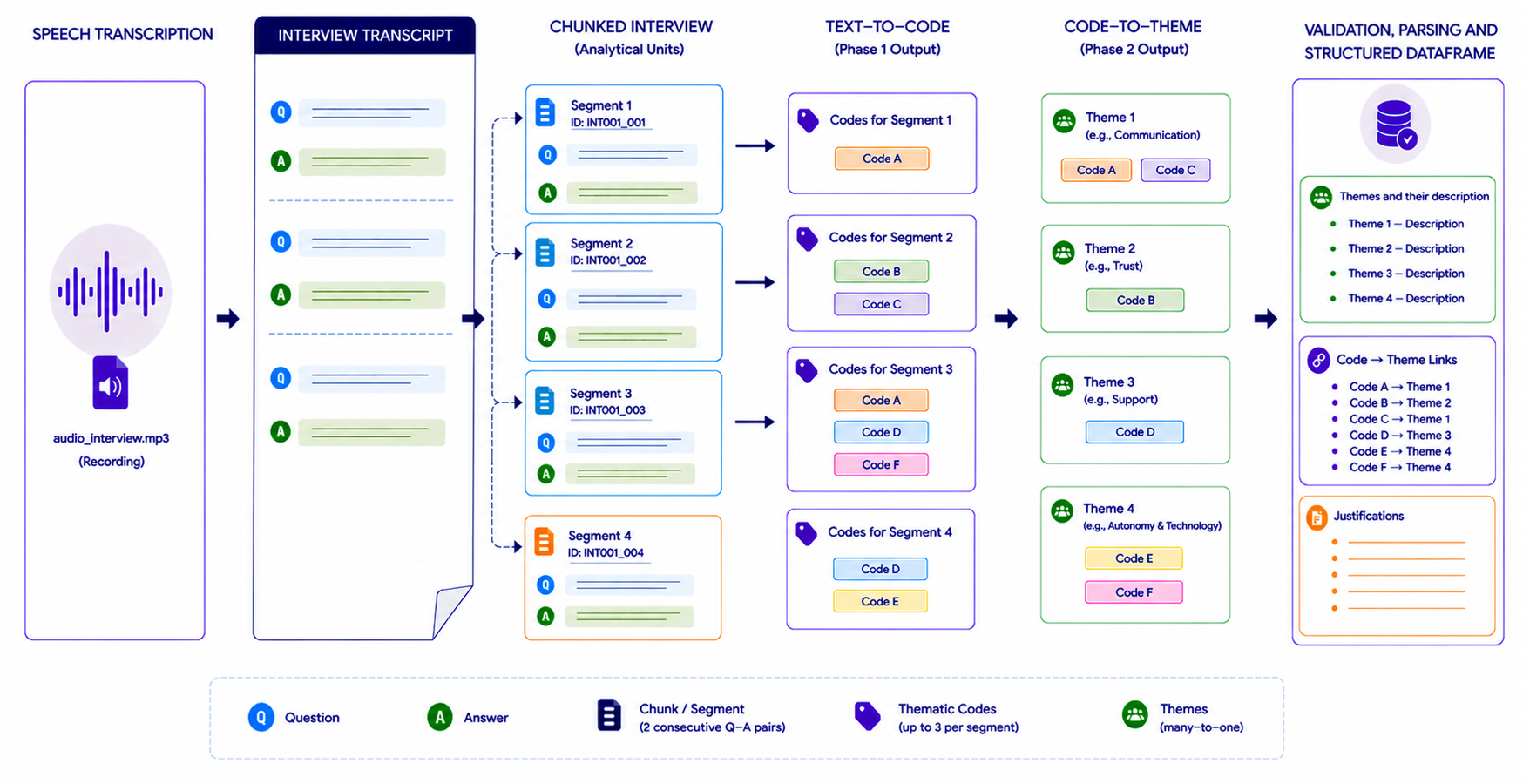}
\caption{Overview of the proposed pipeline. Transcribed interviews are segmented into analytical units(two Q–A pairs). Phase 1 generates segment-level codes with justifications, while Phase 2 aggregates codes into themes via an iterative many-to-one mapping. Outputs are validated and structured to preserve links between segments, codes, themes and justifications.}
\label{fig:ta_pipeline}
\end{figure*} 

Specifically, the\textit{ central interpretative tasks} of generating and representing codes and themes, including the analytical reasoning underlying  the analysis of assignment of codes and themes, are delegated to the LLM through domain-specific prompting strategies. The \textit{procedural task} of ensuring systematic and transparent links between data extracts, codes, and themes are handled programmatically through deterministic processing and validation procedures.

This separation between interpretative inference and procedural control enables the workflow to preserve key methodological requirements of TA while supporting computational traceability and reproducibility. Generated codes and themes remain explicitly linked to their originating data, while all analytical and procedural steps are systematically represented within the workflow (Figure \ref{fig:ta_pipeline}).

A smaller, open-weight model \textit{Mistral 7B v0.3 Instruct} \cite{mistral2024instruct} is used for the interpretative tasks. Recent benchmarking suggests that closely related Mistral models provide functional Danish-language capabilities, although with lower performance compared to larger proprietary models \cite{vejlgaard_holm_danoliteracy_2025} and aligns with Principle 5: Preserve privacy though local deployment.   

Inference was configured deterministically to reduce stochastic variation. Because the workflow consists of analytical tasks with varying output requirements, generation length is constrained to between 40 and 800 new tokens depending on the task.

Figure \ref{fig:ta_pipeline} illustrates the different analytical representations  which  are generated throughout the workflow across Phase 1 and Phase 2:  codes and accompanying justifications, themes and accompanying theme descriptions.

\subsection{Phase 1: From text to codes} 
Each interview segment is processed by the LLM using a three-part, domain-specific prompting strategy inspired by Self-Refine \cite{madaan_self_refine_2023} and which emphazises that each code must be accompanied by an explicit justification (Principle 3). This also reflects the many-to-many relationship between text and codes where a single segment may be assigned multiple codes and similar codes may recur across different segments.  Output generation is constrained through predefined Pydantic schemas \cite{pydantic2024}. These are implemented both during inference using Outlines \cite{outlines2023} and again during parsing and validation to ensure structurally consistent outputs. This step implements principles 1 and 4 by both defining relevant constraining inference. The resulting codes and justifications are subsequently normalized to enforce standardized representations before being integrated into a structured dataframe which link outputs to their corresponding interview segments and identifiers in alignment with Principle 2. Table~\ref{tab:table1} presents an example of outputs from Phase 1.

\begin{table*}[t]
\centering
\footnotesize
\caption{Example of annotated interview segments showing the relationship between text excerpts, assigned codes, and analytical justifications. For readability, the interview segments are here translated to English, although the LLM generated the codes based on the original Danish text.}
\begin{tabular}{ p{6.5cm} p{3.5cm} p{5.2cm}}
\toprule
\textbf{Interview Excerpt} & \textbf{Codes} & \textbf{Justification} \\
\midrule

\textbf{Q: what other jobs have you had?}& 
\multirow[t]{5}{3.5cm}{'\textsc{Work History}', '\textsc{Economic Change}', '\textsc{Positive Life Experience}'} & 
\multirow[t]{5}{5.2cm}{\textbf{i)} Discusses the informant’ s work history and employment in a brewery in Kolding. \textbf{ii)} Mentions the significant increase in wages from DKK (Danish Krone) over time, indicating economic changes in the society during the time period under study. \textbf{iii)} Expresses a positive sentiment about their life, including having a good life and a happy family.}\\
\textbf{A:} \textit{Yes, but then {[the informant’s husband]} and I — Granddad and I —we moved in together and moved to Kolding. And I got a job as a brewery worker at a place called {[brewery]}, which no longer exists. I worked there for…four years, I think, and earned 10 kroner an hour. That was what you earned —10 kroner an hour. Then we got a collective agreement, and our wages went up by 2 kroner. That was an incredible amount of money.}&   &        \\
\\
\textbf{Q: Mm, that was a lot compared with 10 kroner. So…} &   &   \\
\textbf{A:} \textit{Yes, yes. So you should be grateful that you’re able to earn money today. Yes. But, well—and I was actually very happy about that too, and then—yes. A good life. A wonderful life. We got an apartment and then—well, I don’t know, is this where the questions about the children and things like that come in?}&  &    \\

\\
\midrule
\\
\textbf{Q: Yes. The next question is about whether you took maternity leave or parental leave when you had your children.}& 
\multirow[t]{4}{3.5cm}{'\textsc{No Formal Parental Leave}', '\textsc{Duration of Parental Leave}'} & 
\multirow[t]{4}{5.2cm}{\textbf{i)} The informant implies that there was no formal parental leave policy in place during their time as they mention only having three months to be at home with their children without any specific policy being named. \textbf{ii)} The informant specifies a duration of 3 months for parental leave.
}\\
\textbf{A:} \textit{There was no such thing back then.}&   &        \\ \\
\textbf{Q: There was no such thing?} &   &   \\ 
\textbf{A:} \textit{I think we had three months when we could stay at home with our children.}&  &    \\
\bottomrule
\end{tabular}

\label{tab:table1}
\end{table*}

\subsection{Phase 2: From codes to themes}

Codes generated in Phase 1 are used to construct higher-level themes, reflecting a many-to-one relationship between codes and themes, where multiple codes are grouped in a single theme. An initial thematic map is first generated by prompting the LLM on a random batch of 12 codes and corresponding justifications for analytical context\footnote{Different batch sizes were tested but resulted in unstable inference behavior, including hallucinations on the content of the batched codes.}. The remaining codes, along with their associated justifications and interview segments for context, are then processed sequentially. Using a methodological and domain-specific analytical prompting strategy, the LLM assigns each code to an existing theme or mark it as non-applicable.

Unassigned codes are continuously accumulated in a separate pool. When this pool reaches 20 codes, the thematic map is updated by prompting the LLM on a new random subset of 12 codes, allowing for iterative refinement and expansion of the thematic structure. During these updates, previously generated themes are provided to the model as contextual guidance, and the model is instructed to introduce new themes only when the current data contains patterns not already represented in the existing thematic structure. This enables iterative expansion of the thematic map while preserving localized inference at the assignment level. The refinement of the thematic map and assigning codes to themes continues until all codes are either assigned or incorporated through map updates. 

The results of Phase 2 are themes, descriptions of themes as well as  justifications explaining the reasoning for why each code is assigned to its theme, again in alignment with design principles 2-4. Codes that remain unassigned are retained in the pool for documentation, indicating that they may have limited support in the data. As for Phase 1, outputs are constrained, validated, and integrated into a structured dataframe linking themes to their identifiers, thereby ensuring traceability between codes, themes, and interview segments.

Finally, the workflow generates an inverse “codes-in-themes” mapping documenting code-to-theme assignments together with their justifications. Here justifications are provided which explain the analytical reasoning for why codes are assigned to themes. An example of Phase 2 outputs is shown in Table \ref{tab:table2}.  

\renewcommand{\arraystretch}{1.1}
\begin{table*}[t]
\centering
\footnotesize
\caption{Example themes produced during the code-to-theme stage of the thematic analysis pipeline, showing theme descriptions, representative code labels, and the number of codes assigned to each theme.}
\label{tab:table2}
\begin{tabular}{p{3.2cm} p{6cm}  p{4.2cm} p{1.1cm}}
\toprule
\textbf{Theme label }& \textbf{Theme description} & \textbf{Code label} & \textbf{\# Codes} \\ 
\midrule
Struggles   in Education       & The participants' experiences   reveal a shared struggle with education, often characterized by challenges,   repetition, and negative interactions with teachers.                                                   &    '\textsc{economic hardship}', '\textsc{traditional lifestyle}', '\textsc{limited access to child healthcare}', …                     & 10   \\    \\ \midrule \\
Impact   of Feminist Movements & The informants' personal   experiences and direct references to significant feminist movements indicate   a profound impact on their lives, particularly in terms of women's liberation   and changing gender roles. & '\textsc{Gender Role Changes}', '\textsc{supportive attitude}',   '\textsc{non participation in feminist activism of the 80s}',   … & 125   \\   \\ \midrule

\end{tabular}

\end{table*}

Across both phases, the design translates inductive TA into an auditable computational workflow while maintaining core TA requirements such as context sensitivity, data traceability,  and coherent representation of codes and themes.

The  design thus  implements TA in alignment with the central epistemology of the method and as a proof-of-concept. Furthermore, the proof-of-concept supports scalability and transferability across datasets and research contexts, as adaptation primarily involves adjusting the domain-specific prompting strategy to reflect the analytical focus and characteristics of the data. The design in effect also integrates a Privacy By Design approach \cite{srinivasan_start_2018} as local deployment is handled within controlled environments where no data is exposed to external systems.

\section{Evaluation framework}
The proposed design implemements the methodological principles of TA outlines above. However, adherence to the design does not in itself establish whether the resulting codes and themes constitute analytically meaningful interpretations. The central quality criterion of TA - the characteristics of analytically valid outputs (Table \ref{tab:methodological_dimensions}) - therefore must be addressed through evaluation. 

Because inductive TA is inherently interpretative, multiple analytical outputs of the same empirical material may be methodologically valid. Consequently, conventional metrics based on accuracy, semantic similarity or inter-annotator agreement are insufficient as stand-alone measures of analytical quality \cite{beltoft_not_2025, james_counting_2026} . 

We therefore propose an evaluation framework that assesses the computational operationalization from two complementary perspectives: \textbf{(i) }the structural characteristics of the generated analysis and \textbf{(ii)} the analytical quality of its outputs. The first perspective examines how the workflow organizes the empirical material into codes and themes and identifies systematic tendencies relative to human-led analyses. The second perspective examines whether the generated outputs are analytically meaningful in relation to the empirical material and research focus and satisfy output-level quality criteria for TA.

Accordingly, the framework combines a structural comparison with an independent expert assessment. Human-led analyses are used as methodological reference points rather than as a single correct representation of the empirical material.

\subsection{Annotated datasets for evaluation }
To support the evaluation, annotations were produced from the test dataset by three researchers experienced in qualitative analysis. The three annotators worked independently and followed the same methodological procedure. 

One annotator completed a full thematic analysis of the entire test set (n=291 segments), providing a reference for workflow-level evaluation across the two phases. Two additional annotators independently coded a subset of interview segments (n=5) which provided an additional reference only at the coding level. 

\begin{figure}[t]
    \centering
    \includegraphics[width=\columnwidth]{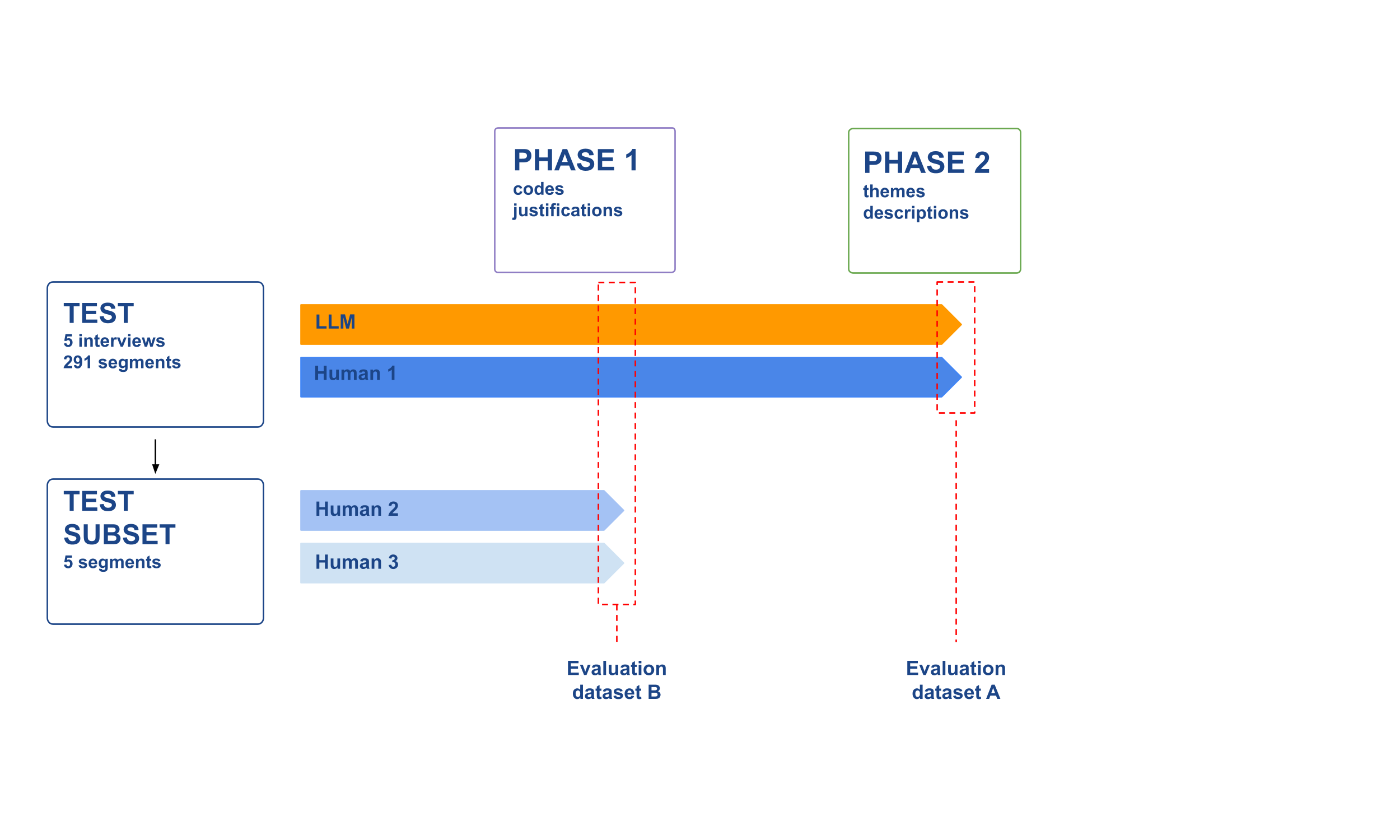}
    \caption{Overview of the evaluation data. The LLM and Human
    Annotator 1 analyzed the full dataset across both Phase 1 and
    Phase 2 (Evaluation Dataset A), while Human Annotators 2 and 3
    independently annotated a smaller subset for the Phase 1
    evaluation only (Evaluation Dataset B).}
    \label{fig:eval_data}
\end{figure}

Together, the two datasets support complementary forms of evaluation. Evaluation Dataset A provides a reference for assessing the complete computational workflow across both coding and theme generation, whereas Evaluation Dataset B supports focused comparison of code-level outputs across multiple human annotators, see Figure \ref{fig:eval_data}.

\subsection{Expert assessment for quality of codes}

To assess the analytical quality of workflow outputs, an expert assessment was conducted using Evaluation Dataset B together with the corresponding outputs generated by the workflow on the same interview excerpts. The assessment was carried out by two independent researchers experienced in qualitative methodology and familiar with the analytical focus of the FREI project.

Outputs were evaluated independently using a purpose-built survey in which annotator outputs were presented in randomized and blinded order to reduce evaluation bias. The evaluation assessed two complementary quality dimensions using five-point Likert scales: \textbf{(i)} code coverage, referring to the extent to which assigned codes captured the meaning of the interview excerpt, and \textbf{(ii)} justification quality, referring to the clarity and adequacy of the analytical reasoning underlying each assigned code.

Beyond its application to the present workflow, the evaluation framework offers a transferable approach for assessing future computational operationalizations of qualitative analytical methods.

Given the relatively small datasets, the present analysis is exploratory and relies primarily on descriptive statistics rather than significance testing. The findings should therefore be interpreted as indicative rather than conclusive. However, the framework can be extended to larger datasets and more comprehensive statistical analyses.

\section{Results}
First, the structural comparison characterizes how the workflow organizes the empirical material into codes and themes relative to human-led analyses, thereby identifying systematic tendencies in the generated analysis. Second, the independent expert assessment examines whether the generated codes and their associated justifications constitute analytically meaningful and well-grounded interpretations of the empirical material. 

\subsection{Structural comparison}

Structural comparisons of the five randomly selected interview segments from evaluation dataset B shows that the LLM in Phase 1 produces a broadly comparable number of codes per segment relative to human annotators. As shown in Table~\ref{tab:code_count_comparison}, the average number of generated codes per segment falls within the range observed across human annotators, although the LLM generally produces slightly fewer codes overall. Across human annotators, variation in the number of codes per segment was also observed.  Given that both the LLM and human annotators were instructed to generate up to three codes per segment, some degree of similarity in coding volume is expected. Nevertheless, the results suggest that the workflow operationalizes Phase 1 at a level broadly comparable to human annotators.

\begin{table*}[t]
\centering
\small

\caption{Number of codes generated by the proposed LLM-based pipeline and three human annotators across five randomly selected interview segments (unit) from the test set (n= 5 segments)}
\label{tab:code_count_comparison}
\vspace{0.3cm}

\begin{tabular}{p{3.8cm}p{2cm} p{1.5cm} p{2cm} p{2cm}}
\toprule
\textbf{Segment id}& \textbf{Human 1}& \textbf{LLM} & \textbf{Human 2}& \textbf{Human 3}\\
\midrule

901  & 1 & 1 & 2 & 2 \\
2333 & 3 & 1 & 1 & 2 \\
4580 & 2 & 1 & 1 & 2 \\
7490 & 1 & 2 & 2 & 2 \\
7524 & 2 & 1 & 2 & 2 \\

\midrule
\textbf{Mean} & \textbf{1.8} & \textbf{1.2} & \textbf{1.6} & \textbf{2.0} \\

\bottomrule
\end{tabular}

\end{table*}

A comparison of average number of code ratios per segment between the LLM workflow and the single annotator in evaluation dataset A shows that the workflow produces an average of 2.33 codes per segment, compared to 1.85 codes per segment for the human annotator. In contrast to the comparison with evaluation dataset B, this indicates that  the LLM workflow performs coding in Phase 1 in a slightly more inclusive manner by assigning a larger number of codes per interview segment. 

However, differences become more apparent when examining code diversity, i.e. how many codes are unique and reused respectively. In TA, code diversity may provide insight into the degree of analytical differentiation and granularity applied during coding, as more diverse code generation can reflect increased sensitivity to variation and nuance within the data. 

Table~\ref{tab:code_diversity_summary} show that the LLM generates fewer unique code labels overall compared to the human annotators, although patterns of code reuse vary across both human and LLM-generated outputs. Given the relatively small absolute differences, the results overall suggest broadly comparable coding behavior in terms of code diversity across human annotators and the workflow.

\begin{table*}[t]
\centering
\small

\caption{Comparison of unique, reused, and total code counts across human annotators and the LLM ($n=5$ segments).}
\label{tab:code_diversity_summary}

\begin{tabular}{p{2cm} p{2cm} p{1.5cm} p{2cm}}
\toprule
\textbf{Annotator} &
\textbf{Unique codes} &
\textbf{Reused codes} &
\textbf{Total codes} \\
\midrule

Human 1 & 9 & 0 & 9 \\
LLM     & 4 & 2 & 6 \\
Human 2 & 8 & 0 & 8 \\
Human 3 & 8 & 2 & 10 \\

\bottomrule
\end{tabular}

\end{table*}

The lower number of unique codes produced by the LLM is particularly noteworthy. This result is interesting given the methodological setup of the workflow, in which the LLM processes each interview segment independently without memory of prior coding decisions. In contrast, human annotators code segments with awareness of previously generated codes and may therefore be expected to reuse existing labels more consistently across segments. From this perspective, it could be expected that the LLM would generate a larger number of unique codes due to the absence of cumulative coding awareness across the dataset. The fact that this is not observed may suggest that the computational workflow is able to meet the quality criteria for "systematic and thorough coding across the entire dataset" (See Table~\ref{tab:methodological_dimensions}) despite the localized inference structure of Phase 1.  

Another plausible explanation for this pattern is the effect of the domain-specific prompting strategy of Phase 1. Given the otherwise high capacity for linguistic variation for the LLM , it is notable that the model does not produce a substantially larger number of unique codes compared to the human annotators. 

The final structural analysis focuses on how the workflow i Phase 2 transforms codes into broader thematic structures relative to human-led TA. 

Table~\ref{tab:comparison_of_code_to_theme} compares overall structural properties of the thematic organization generated by the workflow and the human annotator (evaluation dataset A). The results show substantial structural differences between the thematic composition produced by the workflow and the human annotator. Compared to the human TA, the LLM generates fewer but substantially larger themes in terms of code concentration per theme. This is reflected in both the mean and median number of codes per theme, as well as in the considerably larger maximum theme size produced by the workflow.

\begin{table*}[t]
\centering
\small
\caption{Comparison of theme compositions generated by a human annotator and by the proposed LLM-based pipeline on the test set. Two human-generated themes were merged due to a singular/plural variation in the theme label. The theme description was identical for both labels due to human error (n=291 segments)}
\label{tab:comparison_of_code_to_theme}
\vspace{0.3cm}
\begin{tabular}{p{6cm} p{2cm} p{2cm} }
\toprule
\textbf{Metric}& \textbf{Human}& \textbf{LLM}\\
\midrule

Total number of themes& 19& 10\\
Mean codes per theme& 20.79& 46.7\\
Median codes per theme& 18& 29\\
Largest theme (no. of codes)& 54& 148\\
Smallest theme (no. of codes)& 4& 2\\
\bottomrule
\end{tabular}

\end{table*}

These findings suggest that the workflow operationalizes Phase 2 through a more compressed thematic structure, where higher number of codes are aggregated into broader themes. In contrast, the human annotator produces a more fine-grained thematic composition distributed across a larger number of smaller themes.

\begin{figure}[h]
\centering
\includegraphics[width=0.50\textwidth]{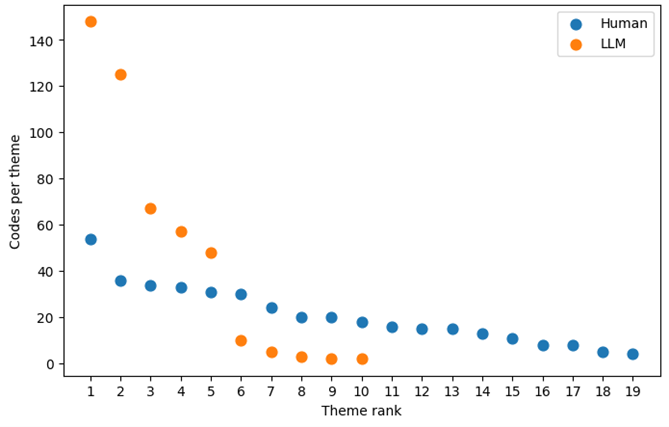}
\caption{Comparison of ranked theme sizes in human- and LLM-generated thematic structures. Distribution of theme sizes measured by the number of assigned codes, ranked from largest to smallest theme (n=291 segments).}
\label{theme_size}
\end{figure}

Figure ~\ref{theme_size} further illustrates differences in how codes are distributed across themes in the human- and LLM-generated thematic compositions. The workflow exhibits a more concentrated thematic composition, where a relatively small number of themes contain a large proportion of the assigned codes. In contrast, the human-generated thematic composition distributes codes more evenly across a larger number of medium-sized themes. 

These differences may partly reflect characteristics of the Phase 2 workflow design, in which codes are iteratively assigned to an evolving thematic map. Because codes are continuously evaluated against already established themes, the workflow may favor incorporation into existing  themes rather than assigning codes into new themes. In contrast, human analysts ideally develop themes with broader contextual awareness of previously generated themes, analytical decisions, and the dataset as a whole. This may  support a  fine-grained, and as such distributed, thematic composition. Although this relationship was not systematically evaluated, the concentration of codes within fewer LLM-generated themes may suggest that the Phase 2 design promotes early thematic consolidation, where initially established themes increasingly shape subsequent code-to-theme assignments. 

\subsection{Expert assessment of quality of codes and justifications }
The expert assessment indicates that the proposed workflow generates code-level outputs comparable to those produced by human annotators, while the accompanying analytical justifications receive consistently higher expert ratings.

\begin{figure*}[ht]
\centering

\begin{minipage}[t]{0.49\textwidth}
    \centering
    \includegraphics[width=\linewidth]{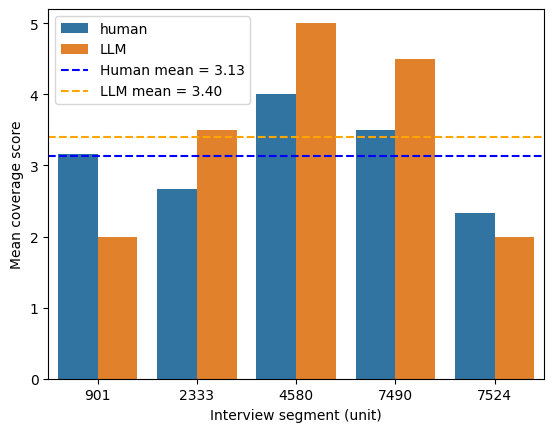}
\end{minipage}
\hfill
\begin{minipage}[t]{0.49\textwidth}
    \centering
    \includegraphics[width=\linewidth]{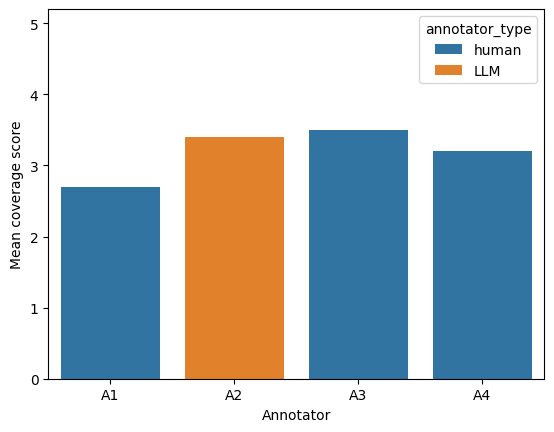}
\end{minipage}

\caption{Comparison of code coverage scores. Left: mean code coverage scores across the five interview segments. Right: overall mean code coverage scores for the human annotators and the proposed LLM. Scores are averaged across the two evaluators ($n=33$).}
\label{fig:coverage_results}
\end{figure*}

Figure~\ref{fig:coverage_results} (left) shows mean code coverage scores between human- and LLM-generated outputs across the five interview segments. Overall, the LLM achieves coverage scores comparable to or higher than the human annotators across several interview segments, although performance varies across cases. The mean coverage score for the LLM is slightly higher overall (3.40) compared to the human annotators (3.13).

The results further show substantial variation between interview segments. In some cases, such as segments 4580 and 7490, the workflow receives noticeably higher coverage scores, whereas the human annotators receive higher scores in segments 901 and 7524. 

Figure ~\ref{fig:coverage_results} (right) zooms in on the scores at the individual annotator level. Here additional variation becomes apparent. The plot shows that mean coverage scores vary substantially across human annotators, with the LLM score falling within the broader range observed across the human evaluations. Notably, the difference between the LLM and individual human annotators is comparable to the variation observed between the human annotators themselves.

These findings suggest that the workflow preserves sufficient contextual information to support meaningful interpretation while enabling the LLM to identify analytically relevant aspects of the empirical material. Despite relying on localized inference where the LLM does not have access to it previous coding decisions, the workflow produces code-level outputs that are broadly comparable to those of human annotators.

\begin{figure*}[ht]
\centering

\begin{minipage}[t]{0.49\textwidth}
    \centering
    \includegraphics[width=\linewidth]{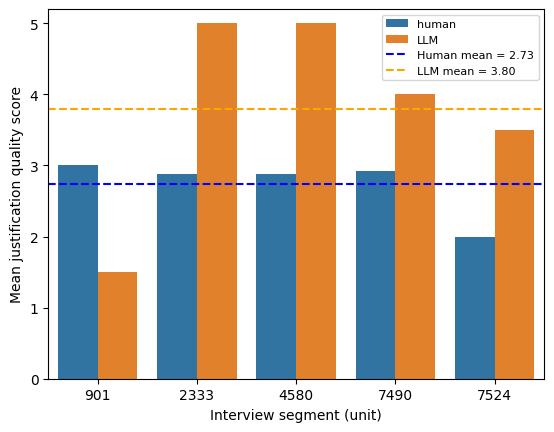}
\end{minipage}
\hfill
\begin{minipage}[t]{0.49\textwidth}
    \centering
    \includegraphics[width=\linewidth]{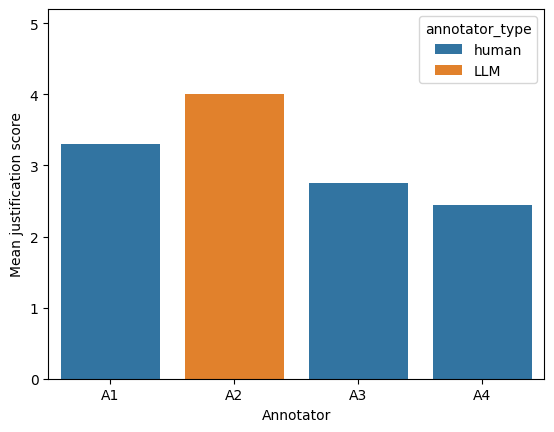}
\end{minipage}

\caption{Comparison of justification quality scores. Left: mean justification quality scores across the five interview segments. Right: overall mean justification quality scores for the human annotators and the proposed LLM. Scores are averaged across the two evaluators ($n=28$).}
\label{fig:justification_results}
\end{figure*}

Turning to the quality of the justifications, Figure ~\ref{fig:justification_results} (right) shows that the workflow generally receives higher justification quality scores across the evaluated interview segments, and in several cases by a substantial margin. Compared to the human annotators, the LLM produces more highly rated analytical justifications in four out of the five interview segments, resulting in a higher overall mean justification score (3.80 compared to 2.73 for the human annotators).

The differences are particularly pronounced in segments 2333 and 4580, where the workflow receives the maximum score, while the human-generated justifications remain closer to the overall human mean. Compared to the code coverage evaluations, the observed pattern is also more consistent across interview segments. This suggests that expert evaluations of justification quality are less sensitive to variation between individual segments.

Figure ~\ref{fig:justification_results} (right) examines the evaluations at the annotator level. The results show that the workflow achieves a higher overall mean justification quality score than all three human annotators. At the same time, substantial variation is also observed across the human annotators themselves, with mean scores ranging from approximately 2.5 to 3.3. Compared to the code coverage quality evaluations, the workflow therefore appears to perform more consistently above the range of the human annotators with regard to justification quality. 

These findings in relation to justification quality suggest that requiring the workflow to generate explicit analytical justifications does not merely improve transparency, but also produces representations of analytical reasoning that experts evaluate as more interpretable than those produced through human annotation.

Although the evaluation data is limited in scope, the results from the expert evaluations of code coverage and justification quality suggest that the workflow produces outputs at a level comparable to human-led TA and in some cases receives higher evaluation scores, particularly in the explicit articulation of analytical reasoning.

\section{Conclusion}
This paper has presented an auditable computational operationalization of inductive and latent thematic analysis. Rather than treating TA as a sequence of generic code- and theme-generation tasks, the proposed approach translates central methodological requirements into five explicit design principles: preserving interpretative context, maintaining traceable relationships between empirical material and analytical outputs, representing analytical constructs and reasoning explicitly, constraining LLM inference to interpretative tasks, and supporting privacy-preserving local deployment.

These principles are operationalized through a functional separation between interpretative LLM inference and deterministic procedural control. While the LLM generates codes, themes, and analytical justifications, programmatic procedures structure, validate, and preserve the relationships between interview segments, codes, and themes. Auditability therefore arises from the workflow architecture as a whole rather than from the transparency of the language model itself.

The exploratory evaluation provides preliminary support for the feasibility of this approach. At the coding level, the workflow produced outputs with expert-rated coverage broadly comparable to that of human annotators. Its analytical justifications received higher mean ratings, indicating that requiring explicit representations of analytical reasoning may strengthen the transparency and interpretability of LLM-supported qualitative analysis. The structural comparison nevertheless showed that the workflow generated fewer and substantially broader themes than the human analysis. This suggests that the sequential Phase 2 procedure favors early thematic consolidation and demonstrates that computational workflow design actively shapes the resulting analytical structure.

These findings should be interpreted cautiously given the limited evaluation sample, the use of a single small open-weight model, and the reliance on one human analysis for the full workflow comparison. Moreover, the analytical quality of the generated themes was not independently assessed. Future work should therefore evaluate the workflow across larger and more diverse datasets, models, annotators, and expert evaluators, while developing Phase 2 procedures that support broader thematic reconsideration and refinement. Overall, the study demonstrates that LLM-supported TA can be computationally operationalized in a methodologically explicit, traceable, and privacy-preserving manner, while also showing that auditability requires systematic attention to the design of the entire analytical workflow


\bibliography{custom}

\appendix

\end{document}